# A NEW 3-DOF 2T1R PARALLEL MECHANISM: TOPOLOGY DESIGN AND KINEMATICS

Huiping Shen[1], Zhongqiu Du[1], Damien Chablat[2], Ju Li[1], Guanglei Wu[3]

[1]Research Center for Advanced Mechanism Theory, Changzhou University, Changzhou 213016, China
[2]Nantes Université, École Centrale Nantes, CNRS, LS2N, UMR 6004, 44000 Nantes, France
[3]School of Mechanical Engineering, Dalian University of Technology, Dalian 116024, China
[1]shp65@126.com, [1]duzq1028@163.com, [2]Damien.Chablat@cnrs.fr, [1]wangju0209@163.com, [3]gwu@dlut.edu.cn

**ABSTRACT**

This article presents a new three-degree-of-freedom (3-DOF) parallel mechanism (PM) with two translations and one rotation (2T1R), designed based on the topological design theory of the parallel mechanism using position and orientation characteristics (POC). The PM is primarily intended for use in package sorting and delivery. The mobile platform of the PM moves along a translation axis, picks up objects from a conveyor belt, and tilts them to either side of the axis. We first calculate the PM's topological characteristics, such as the degree of freedom (DOF) and the degree of coupling, and provide its topological analytical formula to represent the topological information of the PM. Next, we solve the direct and inverse kinematic models based on the kinematic modelling principle using the topological features. The models are purely analytic and are broken down into a series of quadratic equations, making them suitable for use in an industrial robot. We also study the singular configurations to identify the serial and parallel singularities. Using the decoupling properties, we size the mechanism to address the package sorting and depositing problem using an algebraic approach. To determine the smallest segment lengths, we use a cylindrical algebraic decomposition to solve a system with inequalities.

Keywords: parallel mechanism; topological analysis; kinematics; singularity, design

## 1. INTRODUCTION

PMs with fewer degrees of freedom (DOFs) are known for their simple structure, practical motion control, low cost, high stiffness, and high loading capacity. As a result, they have become a popular choice for industrial production [1].

Currently, there is a lot of research and application of parallel mechanisms with three degrees of freedom (3-DOF) and either three translations (3T) or three rotations (3R). For instance, the 3T Tripteron [2], Delta [3-5], and Tsai's 3T [6] PMs have been utilized in 3D printing equipment, while 3R PMs have been employed in the design of "dexterous eye". Additionally, some scholars have proposed 3R PMs for radar tracking, stage lighting, mechanical processing, and other applications.

Studies on 3-DOF 2T1R PMs are relatively limited, but they have the potential for extensive use in devices for grasping, positioning, entertainment, and posture adjustment [1]. Zhang et al. [7] designed a modular 2T1R planar PM, but did not provide kinematic analysis. Qu et al. [8] proposed a decoupled PM synthesis method based on Lie group and screw theory, and obtained a new fully decoupled 2T1R PM in space and plane. Li et al. [9] designed a cooking robot using a 2T1R configuration and analyzed its kinematics, including direct and inverse position analysis, velocity analysis, constant direction workspace, and total workspace analysis. Ding et al. [10] developed a new 2T1R PM based on the theory of POC and analyzed its kinematics. Shen et al. [11] designed and analyzed a decoupled 2T1R PM with symbolic direct position solutions, and evaluated the optimized branch arrangement's impact on kinematics, dynamics, and stiffness performances. Li et al. [12] proposed a novel approach to synthesizing 2R1T and 2T1R 3-DOF redundant drive PMs with a novel closed-loop unit based on Grassmann line geometry and the Atlas method. The PMs obtained using this approach all contain closed-loop units and exhibit good symmetry and comprehensive performance, both in the mechanism and the actuator.

The majority of 2T1R PMs described in the literature lack a symbolic direct kinematic model and often have limited decoupling of movement. These limitations pose challenges for subsequent research, including error analysis, computation of the direct kinematic model, and real-time motion control. [13]

In this article, we employ the topological design theory and method of PM based on POC [14-15] to design a 2T1R PM with zero coupling degree ($\kappa = 0$) and partial kinematic decoupling. We analyze the PM's main topological characteristics, including the DOF and degree of coupling. We also provide calculations for the direct and inverse kinematic models, as well as singularity analysis. The designed 2T1R PM can be used as a delivery platform mechanism for sorting in courier transfer stations.

This article is organized into six sections, as follows. Section 2 presents the case study scenario. In Section 3, a novel 2T1R PM with partial kinematic decoupling is designed using the POC based topological design method of PM. Section 4 describes the kinematic modeling of the designed PM according to the kinematic modeling principle of topological characteristics, and solves its symbolic direct and inverse kinematics. Singular configuration analysis is performed in Section 5. Section 6 introduces a method for defining the lengths based on the decoupling of the PM. The conclusions are presented in the final part.

## 2. CONCEPTUAL DESIGN AND APPLICATION SCENARIOS

The express delivery industry is currently experiencing rapid growth and growing in size every year. As the main hub for express logistics, the transit station requires a large number of human resources. This article considers this issue as the context for the application and aims to reduce the use of human resources. An 2T1R PM is intended to be designed and applied to the hub mechanism for the sorting and delivery of the express transfer station.

The conveyor belt carries the courier object for sorting at speed $v_0$. The 2T1R PM picks up the object at the end of the conveyor belt, identifies the area to be shipped by scanning the object, and then transports them. Once the object is delivered to the designated area, it slides down the slope to the next part of the shipping transport. The schematic diagram of his work surface is depicted in Figure 1.

The goal is to create a 2T1R MP that can move a long distance along the y axis, a short distance along the z axis, and also rotate around the y axis. While a simple serial robot could do this, a parallel robot is preferred because it allows all three motors to move in the same direction at the same time when moving along the y axis. This means that the motors can be made smaller or the system can move more dynamically.

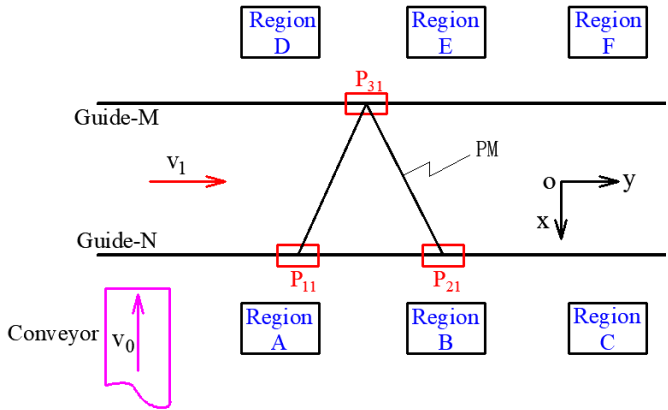

FIGURE 1: CONCEPTUAL DESIGN DIAGRAM OF THE APPLICATION SCENARIO

The schematic diagram of its working plane, the schematic layout of the spatial structure and the three-dimensional drawing of the conceptual design are shown in Figure 2.

## 3. TOPOLOGY DESIGN AND ANALYSIS
### 3.1. Design of the PM

The series and parallel mechanism POC equations are respectively given below [19-21]

$$M_{bi} = \bigcup_{k=1}^{m} M_{J_k} \quad (1)$$

$$M_{pa} = \bigcap_{i=1}^{n} M_{bi} \quad (2)$$

where $M_{J_k}$ - the POC set of the $k$-th kinematic pair, $M_{bi}$ - the set of POC at the end of the $i$-th limb, $M_{Pa}$ - the POC set of the PM moving platform.

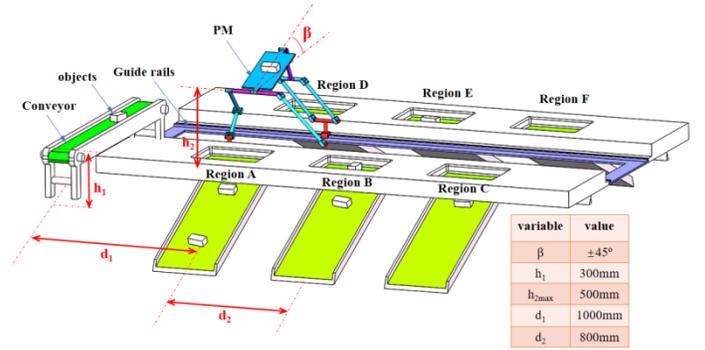

FIGURE 2: CONCEPTUAL DESIGN DIAGRAM OF THE APPLICATION SCENARIO WITH THE DEFINITION DESIGN VARIABLES FIXED.

The object of the article is to design a new partially motion decoupled 2T1R PM. Since the PM containing the hybrid branch chain generally has the advantages of low coupling, input-output (partial) motion decoupling, and easy to solve direct kinematic problem [13, 21].

For this reason, the PM designed in this article is proposed to be completed with two hybrid branched chains. At the same time, considering the sorting, fast delivery operation and dynamic smoothness of the express transfer station, as well as the influence of processing and manufacturing, assembly and other factors, the designed PM mainly adopts the rotary and prismatic pairs to achieve the link constraint between the kinematic limbs.

According to the topological design theory of the PM based on POC [20], it is known from Eq. (2) that in order to realize the moving platform with the motion characteristics of 2T1R, the motion output elements of designed branch chain must contain the motion characteristics of 2T1R.

**1) Design of the hybrid chain** $I$ ($HC\ I$)

First, a spatial sub-PM that can produce two translations (2T) is designed, as shown in

Figure 3(A), which has branch chain A $\{-P_{11}\ ||\ R_{12}\ ||\ R_{13}\ ||\ R_{14}-\}$ and branch chain B $\{-P_{21} \perp R_{22}\ ||\ R_{23}-\}$ located on the same slide rail by moving sliders, respectively, where the symbol "$||$" and "$\perp$" stand for the relationship of parallel and perpendicular for the motion axes of two kinematic pairs, respectively.

For the convenience of elaboration, a coordinate system $o-xyz$ is established on the base platform 0 with the x-axis perpendicular to the direction of the $P_{11}$ axis, the y-axis parallel to the direction of the $P_{11}$ axis, and the z-axis parallel to the direction normal to the base platform 0.

Taking the moving platform with a point on the axis of $R_{24}$ as the base point, the set of POCs of the end links of branch chains A and B are given by Eq. (1) as

$$M_{sub(b1)} = \begin{bmatrix} t^3 \\ r^1(\parallel R_{12}) \end{bmatrix}, \quad M_{sub(b2)} = \begin{bmatrix} t^2(\perp R_{23}) \\ r^1(\parallel R_{22}) \end{bmatrix}$$

Where $t^2(\perp R_{23})$ means that there are two translations lying in a plane normal to the axis of joint $R_{23}$, and, $r^1(\parallel R_{22})$ means that there is one rotation with the axis of rotation parallel to the axis of joint $R_{22}$. The other notations in the formulas above can be found in [14, 15].

From Eq. (2), the set of POCs at the output of the end of the sub-PM is

$$M_{b1} = \begin{bmatrix} t^3 \\ r^1(\parallel R_{12}) \end{bmatrix} \cap \begin{bmatrix} t^2(\perp R_{23}) \\ r^1(\parallel R_{22}) \end{bmatrix} = \begin{bmatrix} t^2(\parallel \Diamond(yoz)) \\ r^0 \end{bmatrix}$$

Thus, the sub-PM is capable of producing a two-dimensional translation motion parallel to the $yoz$ plane, noted as 2T.

To achieve another rotating output, from Eq. (1), a rotary pair $R_{24}$, which is co-axial with the rotary pair $R_{14}$, can be directly connected in series with the output link of the sub-PM, such that the HC *I* is obtained, as shown in Figure 3(B).

**2) Design of the hybrid chain *II* (*HC II*)**

Similarly, when designing the *HC II*, at least the kinematic characteristics of 2T1R should be realized. Here a parallelogram mechanism (noted π joint) consisting of four rotary pairs with good loading capacity and stiffness is used for the design.

First, as shown in Figure 3(C)-(D), at the other side of the base platform 0 slide, one short side of the π joint is rigidly connected to the prismatic pair $P_{31}$, two rotary pairs $R_{33}$ and $R_{34}$ are connected in series on the another short side of the parallelogram mechanism. From Eq. (2), in order to achieve the same rotational characteristics as the rotary pairs $R_{24}$, the axes of the rotary pairs $R_{33}$, $R_{34}$ and $R_{24}$ must be arranged parallel to each other. Therefore, the whole *HC II* can be designed and noted as $\{-P_{31}-P_a-R_{33}\parallel R_{34}-\}$.

From Eq. (1), the sub-chain of the prismatic pair $P_{31}$ rigidly connected to one short side of the π joint and the set of POCs of the sub-chain $R_{33}\parallel R_{34}$ are

$$M_{2p} = \begin{bmatrix} t^1(\parallel P_{31}) \cup t^1(\parallel \pi) \\ r^0 \end{bmatrix}, \quad M_{2R} = \begin{bmatrix} t^2(\parallel \Diamond(xoz)) \\ r^1(\parallel R_{33}) \end{bmatrix}$$

Then the set of POCs of the end members of the *HC II* is

$$M_{b2} = M_{2p} \cup M_{2R} = \begin{bmatrix} t^3 \\ r^1(\parallel R_{33}) \end{bmatrix}$$

Thus, from Eq. (2), the POC set of the moving platform of the PM can be calculated as.

$$M_{pa} = M_{b1} \cap M_{b2} = \begin{bmatrix} t^2(\perp R_{23}) \\ r^1(\parallel R_{24}) \end{bmatrix} \cap \begin{bmatrix} t^3 \\ r^1(\parallel R_{33}) \end{bmatrix} = \begin{bmatrix} t^2(\perp R_{23}) \\ r^1(\parallel R_{24}) \end{bmatrix} \quad (3)$$

From Eq. (3), it can be seen that the moving platform 1 can produce two-dimensional translation motion (Y, Z) in the plane of $yoz$ and one-dimensional rotation around the axis of $R_{24}$. At this point, the design of the 2T1R PM is fully completed. This PM can be used as a sorting and delivery platform device for transfer stations.

### 3.2. Topological analysis
**1) DOF calculation**

The equation for the non-instantaneous DOF (also known as the full circumference DOF) is [19-21]

$$F = \sum_{i=1}^{m} f_i - \sum_{j=1}^{v} \xi_{L_j} \quad (4)$$

$$\xi_{Lj} = \dim.\left\{\left(\bigcap_{i=1}^{j} M_{b_i}\right) \cup M_{b_{(j+1)}}\right\} \quad (5)$$

Where
- $F$ is the DOF of the PM,
- $f_i$ is the DOF of the *i-th* motion pair (without local DOF),
- $m$ is the number of motion pairs contained in the PM,
- $v$ is the number of independent loops ($v = m - n + 1$,

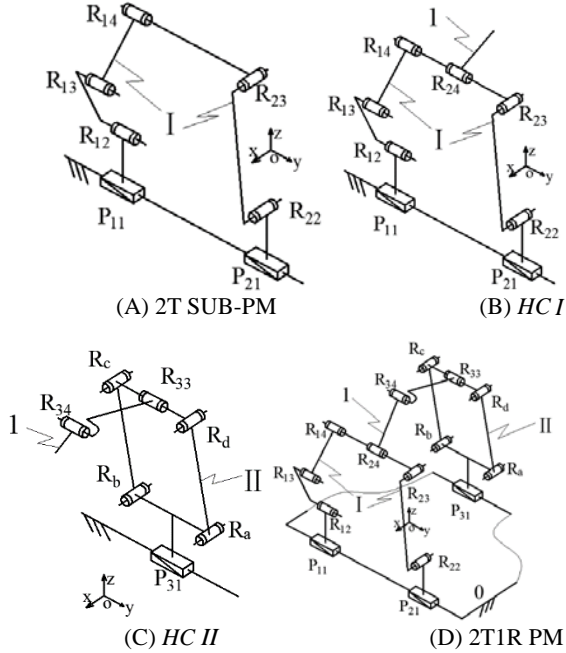

(A) 2T SUB-PM  (B) *HC I*

(C) *HC II*  (D) 2T1R PM

FIGURE 3: DESIGN OF THE 2T1R PM WITH THE HYBRID CHAINS I AND II

© 2023 by ASME

- $n$ is the number of links contained in the PM),
- $\xi_{L_j}$ is the number of independent equations for the $j$-th loop,
- $\bigcap_{i=1}^{j} M_{b_i}$ is the set of *POC* for the sub-PM consisting of the $j$ branched limbs,
- $M_{b(j+1)}$ is the set of *POC* for the end member of the $(j+1)$-th branched limb.

The first loop (i.e., 1st sub-PM) is selected as the above described spatial sub-PM with 2T, noted as
$$loop_1 \{-P_{11} \parallel R_{12} \parallel R_{13} \parallel R_{14} \perp R_{23} \parallel R_{22} \perp P_{21}-\}.$$

Its independent displacement equation number $\xi_{L1}$ is calculated from Eq. (5) as

$$\xi_{L1} = \dim.\left\{ \begin{bmatrix} t^3 \\ r^1(\parallel R_{12}) \end{bmatrix} \cup \begin{bmatrix} t^2(\perp R_{23}) \\ r^1(\parallel R_{22}) \end{bmatrix} \right\}$$

$$= \dim\left\{ \begin{bmatrix} t^3 \\ r^2(\parallel \Diamond(R_{12}, R_{22})) \end{bmatrix} \right\} = 5$$

The DOF of this sub-PM are given by Eq. (4)

$$F'_{(1-2)} = \sum_{i=1}^{m} f_i - \sum_{j=1}^{1} \xi_{Lj} = 7 - 5 = 2$$

The 2nd loop consists of the above sub-PM, the rotary pair $R_{24}$ and limb $\{-P_{31} - P_a - R_{33} \parallel R_{34} -\}$, noted as
$$loop_2 \{-P_{31} - P_a - R_{33} \parallel R_{34} - R_{24} -\}.$$

The number of its independent displacement equation $\xi_{L2}$ is calculated from Eq. (5) as

$$\xi_{L2} = \dim.\left\{ \begin{bmatrix} t^2(\perp R_{23}) \\ r^1(\parallel R_{24}) \end{bmatrix} \cup \begin{bmatrix} t^3 \\ r^1(\parallel R_{33}) \end{bmatrix} \right\} = \dim\left\{ \begin{bmatrix} t^3 \\ r^1(\parallel R_{24}) \end{bmatrix} \right\} = 4$$

Therefore, the DOF of the PM are calculated from Eq. (4) as
$$F = \sum_{i=1}^{m} f_i - \sum_{j=1}^{2} \xi_{Lj} = (8+4)-(5+4) = 3$$

That is, when the prismatic pairs $P_{11}$, $P_{21}$, $P_{31}$ on the base platform 0 are taken as the actuated pairs, the moving platform 1 can realize the motion output of two-dimensional translations in the $yoz$ plane and one-dimensional rotation around the y-axis.

**2) Calculation of the coupling degree**

Based on the composition principle of PM based on single-open-chain ($SOC$) unit [15] it is known that any PM can be decomposed into three types of the ordered $SOC$s with positive, zero and negative constraint degrees, the constraint degree of the $j-th$ $SOC_j$ is defined as [14-16]

$$\Delta_j = \sum_{i=1}^{m_j} f_i - I_j - \xi_{L_j} = \begin{cases} \Delta_j^- = -5, -4, -3, -2, -1 \\ \Delta_j^0 = 0 \\ \Delta_j^+ = +1, +2, +3 \end{cases} \quad (6)$$

Where, $m_j$ is the number of motion pairs for the $j$-th $SOC_j$, $I_j$ is the number of actuated pairs for the $j$-th $SOC_j$.

One ordered set of $v-th$ $SOC$ can be divided into a number of minimal sub-kinematics chains (Sub-kinematics chains, SKC), while each $SKC$ containing only one basic kinematic chain ($BKC$) with zero DOF. For a $SKC$ must satisfy the following formula [14-16].

$$\sum_{j=1}^{v} \Delta_j = 0$$

Thus, the coupling degree of $SKC$ is defined as [14-16]

$$\kappa = \frac{1}{2}\min\left(\sum_{j=1}^{v} \Delta_j \right) \quad (7)$$

Where, $\kappa$ is the coupling degree of $SKC$, $\min(\cdot)$ is the decomposition of $\sum |\Delta_j|$ that is the smallest among the multiple schemes for a $SKC$ to decompose into the ordered $v$-th $SOC(\Delta_j)$.

The constraint degree for each of the two loops of the PM is calculated from Eq. (6) as

$$\Delta_1 = \sum_{i=1}^{m} f_i - I_1 - \xi_{L1} = 7 - 2 - 5 = 0$$

$$\Delta_2 = \sum_{i=1}^{m} f_i - I_2 - \xi_{L2} = 5 - 1 - 4 = 0$$

From the criterion for determining SKCs [21], the PM has 2 SKCs with same coupling degree that are both equal to

$$\kappa = \frac{1}{2}\min\left(\sum_{j=1}^{v} |\Delta_j| \right) = \frac{1}{2} \times 0 = 0$$

**3) Motion decoupling analysis and SKC-based PM topological formulation**

The topological formulation of the PM can be analytically expressed as [16]

$$^{2T1R}PM^0\left[3, 2(7,5)\right] = {}^{2T1R}SKC_1^0(0;5) + {}^{3T1R}SKC_2^0(0;4) \quad (8)$$

Eq. (8) shows that the position analysis of the PM can be transformed into the position analysis of 2 SKCs. Because of the coupling degree $\kappa = 0$, its direct kinematic problem can be solved directly without setting dummy variables [16].

From Eq.(8), it can be seen that the PM contains two $SKCs$ and three actuated prismatic pairs $P_{11}$, $P_{21}$ and $P_{31}$ are distributed in these two $SKCs$. It is easy to know that: in $SKC_1$, the position component of the base point of the moving platform $x$ in the axial direction is constant. While the position component of $y$ in the axial direction is determined only by the prismatic pair $P_{11}$. Since the two-dimensional translation movement generated by the base point of the moving platform is in the $yoz$ plane, the position component of $z$ in the axial direction is determined jointly by $P_{11}$ and $P_{21}$. Thus, the PM has $I-O$ partial motion decoupling property.

## 4. DIRECT AND INVERSE KINEMATIC MODELS

The kinematic modeling of the PM is shown in Figure 4(A). The distance between the two parallel guide rails of the base platform 0 is set as $2a$, and the origin of the base coordinate system is located at a fixed point on the bisector of the parallel lines of the two guide rails. At the same time, set $A_iB_i = l_1 (i=1,2,3)$, $B_1C_1 = C_1D = l_2$, $DE = EC_2 = l_3$, $B_2C_2 = l_4$, $B_3C_3 = l_5$, $C_3F = l_6$, $EF = l_7$ and the rotating angle of the moving platform is $\beta$.

Take the base point E on the moving platform 1 as the origin $O'$ of the moving coordinate system, $x$ and $y$ are respectively vertical and coincide with $DC_2$, and the $z$ axis is determined by the right-handed screw rule.

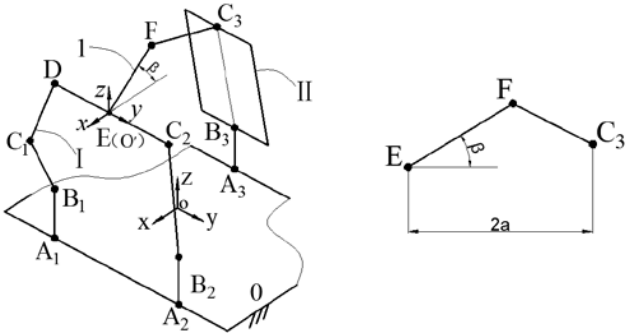

(A) 2T1R PM KINEMATIC MODELING  (B) PROJECTION OF E-F-C_3 ON THE FACE OF $xoz$

FIGURE 4: KINEMATIC MODELING OF THE PM

### 4.1. Direct kinematic model

The direct kinematic model depends on the actuated prismatic joint. So, $y_1$, $y_2$, $y_3$ are known, and the position $O'(x,y,z)$ and the Euler angle $\beta$ of position of the moving platform 1 need to be found. The solutions are called the assembly modes.

The direct kinematic problem of the PM can be solved independently in $SKC_1$ and $SKC_2$, respectively.

**1) Direct kinematics of $SKC_1$**

From Figure 4, we have
$A_1 = (a, y_1, 0)$, $A_2 = (a, y_2, 0)$, $B_1 = (a, y_1, l_1)$,
$B_2 = (a, y_2, l_1)$, $D = (a, y_1, z)$, $C_2 = (a, y_1 + 2l_3, z)$,
$E = O' = (a, y_1 + l_3, z)$.

It can be known from Eq. (3) that
$$y = y_1 + l_3 \tag{9}$$

From the link length constraint $B_2C_2 = l_4$, we get
$$z = l_1 + m\sqrt{l_4^2 - (y_1 + 2l_3 - y_2)^2}, (m = \pm 1) \tag{10}$$

**2) Direct kinematics of $SKC_2$**

It is easy to know $A_3 = (-a, y_3, 0)$ and $B_3 = (-a, y_3, l_1)$.

The projection of the $E-F-C_3$ in this PM on the $xoz$ plane is shown in Figure 4(B). Obviously, we have
$$F = (a - l_7\cos\beta, y_1 + l_3, z + l_7\sin\beta),$$
$$C_3 = (-a, y_1 + l_3, z_{c3})$$

From the link length constraint $B_3C_3 = l_5$, we can get
$$z_{c3} = l_1 + n\sqrt{l_5^2 - (y_1 + l_3 - y_3)^2}, \text{with } (n = \pm 1) \tag{11}$$

And by the link length constraint $C_3F = l_6$, we can get
$$\beta = 2\arctan\frac{A + q\sqrt{A^2 + B^2 - C^2}}{B+C}, \text{with } (q = \pm 1) \tag{12}$$

Where $A = 2(z - z_{c3})l_7$, $B = -4al_7$, $C = l_6^2 - 4a^2 - l_7^2 - (z - z_{c3})^2$.

At this point, the position $O'(x,y,z)$ and rotating angle $\beta$ of the moving platform 1 have been solved. There are
$$\begin{cases} y = f_1(y_1) \\ z = f_2(y_1, y_2) \\ \beta = f_3(y_1, y_2, y_3) \end{cases} \tag{13}$$

From equation (13), it is evident that the 2T1R PM possesses not only a symbolic solution for direct kinematics, but also partial motion decoupling. This means that the platform's positions and orientation are obtained independently from the joint positions. That is, the $y$ value of the base point $O'$ depends on the input value $y_1$, and the $z$ value depends on $y_1$ and $y_2$. Then, $\beta$ depends on $y_1$, $y_2$ and $y_3$. In this way, the results analyzed by Eq. (13) are completely consistent with the results of the $SKC$ based PM topology analysis and its motion decoupling analysis.

Due to $m = \pm 1$ in Eq.(10), $n = \pm 1$ in Eq.(11) and $q = \pm 1$ in Eq.(12), the number of assembly modes is height.

### 4.2. Inverse kinematic models

The inverse solution of the position is that the position $O'(x,y,z)$ and the rotating angle $\beta$ of the moving platform 1 are known. The input $y_1$, $y_2$ and $y_3$ of the actuated prismatic joints have to be found. The solutions are called the working modes.

From $E = (a, y_1 + l_3, z)$, it can be known from Eq.(9) that
$$y_1 = y - l_3 \tag{14}$$

From Eq. (10), it is easy to know
$$y_2 = y + l_3 + u\sqrt{l_4^2 - (z - l_1)^2}, (u = \pm 1) \tag{15}$$

From Eq. (11), it is easy to know
$$y_3 = y + v\sqrt{l_5^2 - (z_{c3} - l_1)^2}, (v = \pm 1) \tag{16}$$

Among them, the link length constraint $C_3F = l_6$ can be obtained as
$$z_{c3} = z + l_7\sin\beta + w\sqrt{l_6^2 - (2a - l_7\cos\beta)^2}, (w = \pm 1) \tag{17}$$

At this point, the inverse kinematics of the PM, ie finding $y_1$, $y_2$, $y_3$ is solved. Therefore, when the base point $O'(x,y,z)$

on the moving platform is known, since $u = \pm 1$ in Eq. (15), $v = \pm 1$ in Eq. (16), and $w = \pm 1$ in Eq. (17), there are two different sets of solutions for the input value $y_i (i = 2,3)$, $z_{c3}$ also has two sets of solutions. Therefore, the number of inverse kinematics solutions of the PM is also height.

### 4.3. Numerical verification of direct and inverse kinematics

Let the dimension parameters of the PM be $a = 300$, $l_1 = 100$, $l_2 = 200$, $l_3 = 160$, $l_4 = 400$, $l_5 = 320$, $l_6 = 240$, and $l_7 = 500$ (unit: mm).

**1) Numerical verification of direct kinematics**

According to the above dimensional parameters, the CAD model of the PM is shown in Figure 5. The input values of the three driving pairs $P_{11}$, $P_{21}$ and $P_{31}$ measured from the 3D model are $y_1 = -244.59$, $y_2 = 303.32$ and $y_3 = -252.26$, respectively, and the output values of the corresponding point $O'$ are $y = -84.59$, $z = 428.72$, and $\beta = 0.3045 (rad)$, respectively.

By using Matlab software, by substituting the input values $y_1 = -244.59$, $y_2 = 303.32$ and $y_3 = -252.26$ into the direct kinematic model from Eqs. (9)-(12), we obtain height direct kinematics solutions, as shown in Table 1. Among them, No.3-6 are imaginary solutions. Therefore, the PM only contains four real solutions, which corresponds to four assembly modes, as shown in Figure 5.

TABLE 1: ASSEMBLY MODES FOR $y_1 = -244.59$, $y_2 = 303.32$ AND $y_3 = -252.26$

| # | $y(mm)$ | $z(mm)$ | $\beta(rad)$ |
|---|---|---|---|
| 1* | -84.5900 | 428.7203 | 0.3045 |
| 2 | -84.5900 | 428.7203 | -0.4912 |
| 3 | -84.5900 | 428.7203 | -0.7865+ 0.3873i |
| 4 | -84.5900 | 428.7203 | -0.7865- 0.3873i |
| 5 | -84.5900 | -228.7203 | 0.7865+ 0.3873i |
| 6 | -84.5900 | -228.7203 | 0.7865- 0.3873i |
| 7 | -84.5900 | -228.7203 | 0.4912 |
| 8 | -84.5900 | -228.7203 | -0.3045 |

Taking into account practical application, we remove three configurations where link interference occurs. The configuration in Figure 5 (A) corresponding to solution #1 in Table 1 is depicted. This configuration corresponds to $m = +1$ in Eq. (10), $n = +1$ in Eq. (11) and $q = -1$ in Eq. (12).

**2) Numerical verification of inverse kinematics**

By substituting the output values $y = -84.59$, $z = 428.72$ and $\beta = 0.3045 (rad)$ of the base point $O'$ of the moving platform measured in the 3D model into Eqs. (14)-(17), we obtain the corresponding height working modes, as shown in Table 2. Among which, solution 1, 2, 5 and 6 are imaginary solutions. Therefore, there are four corresponding working modes, as shown in Figure 6. By combining Table 2 with Figure 6(B), it is apparent that the fourth assembly modes corresponds to the working mode in Figure 5(A).

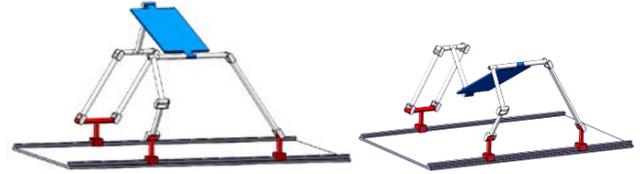

(A) $No.1(m = +1, n = +1, q = -1)$ (B) $No.2(m = +1, n = +1, q = +1)$

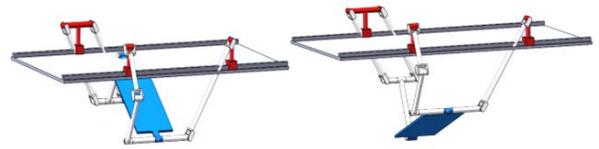

(C) $No.7(m = -1, n = -1, q = -1)$ (D) $No.8(m = -1, n = -1, q = +1)$

FIGURE 5: THE FOUR ASSEMBLY MODES

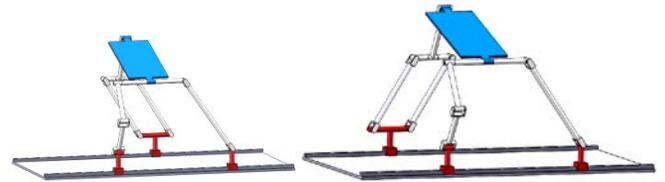

(A) $No.3(u = +1, v = +1, w = -1)$  (B) $No.4(u = +1, v = -1, w = -1)$

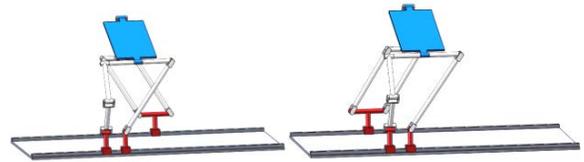

(C) $No.7(u = -1, v = +1, w = -1)$  (D) $No.8(u = -1, v = -1, w = -1)$

FIGURE 6: THE FOUR WORKING MODES

TABLE 2: WORKING MODES FOR $y = -84.59$, $z = 428.72$ AND $\beta = 0.3045 (rad)$

| # | $y_1 (mm)$ | $y_2 (mm)$ | $y_3 (mm)$ |
|---|---|---|---|
| 1 | -244.5900 | 303.3200 | -84.5900+ 605.3355i |
| 2 | -244.5900 | 303.3200 | -84.5900- 605.3355i |
| 3 | -244.5900 | 303.3200 | 83.0989 |
| 4* | -244.5900 | 303.3200 | -252.2789 |
| 5 | -244.5900 | -152.500 | -84.5900+ 605.3355i |
| 6 | -244.5900 | -152.500 | -84.5900- 605.3355i |
| 7 | -244.5900 | -152.500 | 83.0989 |
| 8 | -244.5900 | -152.500 | -252.2789 |

 

## 5. SINGULARITY ANALYSIS

This paper focuses on the study of Jacobian matrices, which establish a relationship between the input speeds associated with the actuated joints and the speed of the mobile platform [16-21]. Using the constraint equations that link the inputs and outputs of the PM and differentiate them from time, we obtain the following corresponding relation:

$$\mathbf{A}\mathbf{t} + \mathbf{B}\dot{\rho} = 0 \qquad (18)$$

Where, $\mathbf{A}$ and $\mathbf{B}$ are respectively the parallel and serial Jacobian matrices, $\mathbf{t} = \begin{bmatrix} \dot{y} & \dot{z} & \dot{\beta} \end{bmatrix}^T$ is the velocity of the mobile platform and $\dot{\rho} = \begin{bmatrix} \dot{y}_1 & \dot{y}_2 & \dot{y}_3 \end{bmatrix}^T$ the joint velocities. A singularity occurs whenever $\mathbf{A}$ or $\mathbf{B}$, (or both) that can no longer be inverted. Three types of singularities exist [18]:

- $\det(\mathbf{A}) = 0$, a parallel singularity occurs,
- $\det(\mathbf{B}) = 0$, a serial singularity occurs,
- $\det(\mathbf{A}) = 0$ and $\det(\mathbf{B}) = 0$.

There are also constraint singularities that are not studied in this article [21]. We have

$$\mathbf{A} = \begin{bmatrix} f_{11} & f_{12} & f_{13} \\ f_{21} & f_{22} & f_{23} \\ f_{31} & f_{32} & f_{33} \end{bmatrix}, \quad \mathbf{B} = \begin{bmatrix} g_{11} & 0 & 0 \\ 0 & g_{22} & 0 \\ 0 & 0 & g_{33} \end{bmatrix}$$

$$f_{11} = \frac{\partial f_1}{\partial y} = -2(y_1 - y), \quad f_{12} = \frac{\partial f_1}{\partial z} = 0, \quad f_{13} = \frac{\partial f_1}{\partial \beta} = 0,$$

$$f_{21} = \frac{\partial f_2}{\partial y} = 2(y + l_3 - y_2), \quad f_{22} = \frac{\partial f_2}{\partial z} = 2(z - l_1), \quad f_{23} = \frac{\partial f_2}{\partial \beta} = 0,$$

$$f_{31} = \frac{\partial f_3}{\partial y} = \frac{2(z_{c3} - z_F)(y_3 - y)}{\sqrt{l_5^2 - (y - y_3)^2}}, \quad f_{32} = \frac{\partial f_3}{\partial z} = -2(z_{c3} - z_F),$$

$$f_{33} = \frac{\partial f_3}{\partial \beta} = -2l_7 \sin\beta(-2a + l_7 \cos\beta) - 2l_7 \cos\beta(z_{c3} - z_F),$$

$$g_{11} = \frac{\partial f_1}{\partial y_1} = 2(y_1 - y), \quad g_{22} = \frac{\partial f_2}{\partial y_2} = 2(y_2 - y - l_3),$$

$$g_{33} = \frac{\partial f_3}{\partial y_3} = \frac{2(z_{c3} - z_F)(y - y_3)}{\sqrt{l_5^2 - (y - y_3)^2}}$$

**1) Serial singularities**

When $\det(\mathbf{B}) = 0$, the PM undergoes an serial singularity, in the three cases below.

a) $g_{11} = \frac{\partial f_1}{\partial y_1} = 2(y_1 - y) = 0$, it is clear that the case does not satisfy the conformation requirement.

b) $g_{22} = \frac{\partial f_2}{\partial y_2} = 2(y_2 - y - l_3) = 0$, i.e. the coordinates of the y-axis of $A_2$ and $C_2$ are equal, as in Figure 7(A).

c) $g_{33} = \frac{\partial f_3}{\partial y_3} = \frac{2(z_{c3} - z_F)(y - y_3)}{\sqrt{l_5^2 - (y - y_3)^2}} = 0$, when the serial singularity occurs when the coordinates of the z-axis of $C_3$ and $F$ are equal or the coordinates of the y-axis of $E$ and $A_3$ are equal, as shown in Figure 7(B) and Figure 7(C).

When a serial singularity appears, two working modes come together. Serial singularities also constitute the boundaries of the workspace.

We have another singularity when $B_1$, $C_1$ and $D$ are aligned but we cannot detect by the study of the Jacobian matrix because there is no actuated joint associated [23].

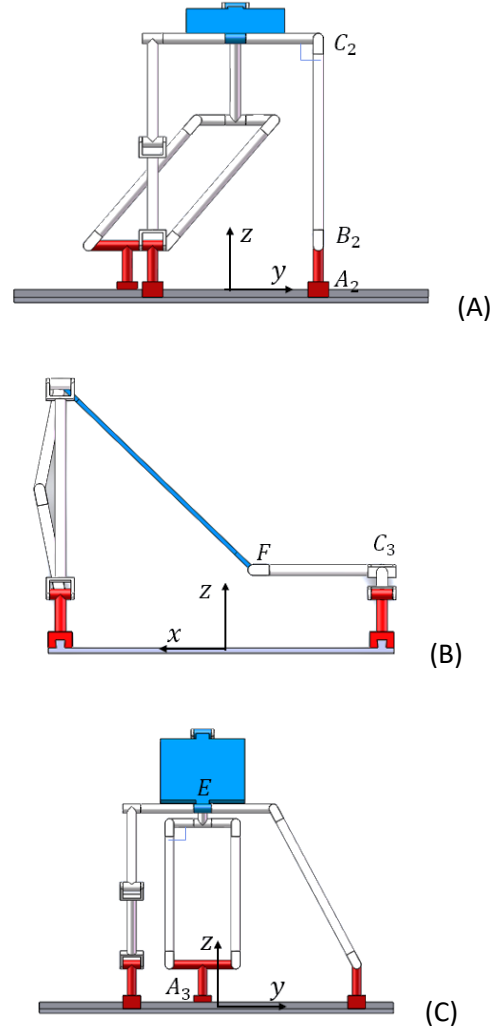

FIGURE 7: SERIAL SINGULARITIES

**2) Parallel singularities**

The parallel singularity of the PM occurs when $\det(\mathbf{A}) = 0$. Since $\mathbf{A}$ is a lower triangular matrix, there are also three cases of output singularities.

a) $f_{11} = \frac{\partial f_1}{\partial y} = 0$, it is when the platform changes its assembly mode between the bottom and the top.

b) $f_{22} = \frac{\partial f_2}{\partial z} = 0$, at which point the matrix **A** is not full rank and the output singularity occurs when $\det(\mathbf{A}) = 0$, i.e., the z-axis coordinates of the point $E$ and the point $B_2$ are equal, and the singular position does not exist in this case because the interference between the link lengths does not reach that position.

c) $f_{33} = \frac{\partial f_3}{\partial \beta} = 0$, i.e., the output singularity of the PM occurs when the Euler angle of the moving platform satisfies $\beta = \arctan\left(\frac{z_{c3} - z}{2a}\right)$.

## 6. DEFINITION OF ROBOT'S LENGTHS TO MEET THE PROPOSED SCENARIO

The aim of this section is to determine the dimensions of the robot that satisfy the requirements outlined in Section 2. Thanks to the decoupling properties, these calculations can be carried out with ease.

**1) Dimensioning of the mobile platform.**

We have to define $l_6$ and $l_7$ to achieve +/-45° travel to drop packages from the treadmill (0°) to the shipping boxes. We can write

$$(-600 + l_7 \cos(\beta))^2 + (\rho_1 + l_7 \sin(\beta) - \rho_2)^2 = l_6^2 \quad (19)$$

Where $\rho_1$ and $\rho_2$ are the motion along z provide by the two opposite loop. The distance between the hoppers to deposit the objects being fixed ($2a$), $l_6$ is defined by β, for +/- 45° where $l_7 = 2a\sqrt{2}$.

To avoid the parallel singularity, we define $\theta$ angle between bars (EF) and (FC3). As the singularity is when $\theta = 0 \pm \pi$, we set $\theta \geq 2/10$. We have

$$\sin(\theta) l_6 l_7 = (\rho_1 + l_7 \sin(\beta) - \rho_2) l_7 \cos(\beta) + (600 - l_7 \cos(\beta)) l_7 \sin(\beta) \quad (20)$$

$$\cos(\theta) l_6 l_7 = (600 - l_7 \cos(\beta)) l_7 \cos(\beta) - (\rho_1 + l_7 \sin(\beta) - \rho_2) l_7 \sin(\beta) \quad (21)$$

We are looking for all $l_7$ allowing a movement with $\theta \geq 2/10$.

$$l_6^2 = (-600 + 600\sqrt{2} \cos(\beta))^2 + (600\sqrt{2} \sin(\beta) - \rho_2)^2$$

$$0 = 600 \cos(\theta) l_6 \sqrt{2} + 600(-600 + 600\sqrt{2} \cos(\beta))\sqrt{2} \cos(\beta) + 600(600\sqrt{2} \sin(\beta) - \rho_2)\sqrt{2} \sin(\beta)$$

With some inequalities

$$\sin(\theta) \geq \frac{1}{5}$$
$$\sin(\beta) \geq -\frac{\sqrt{2}}{2}$$
$$\sin(\beta) \leq \frac{\sqrt{2}}{2} \quad (22)$$
$$l_6 \geq 0$$

From Figure 8, when β= 45° and β= -45°, platform is at the level of the package evacuation guides while for β= 0°, it is at the level of the loading conveyor belt which is a height $h_1$.

The methodology used was introduced in [23], to solve the similar problem. The methodology involves the use of Cylindrical Algebraic Decomposition (CAD), where cells are bounded by the discriminating variety [24]. In our case, when the CAD is calculated based on $\beta$, we have 11 equations.

Two equations have a similar role in defining the smallest value of $l_6$. We study the following equation to find its maximum which will become the minimum value of $l_6$ to represent the motion without singularity.

$$36\left(\tan^8\left(\frac{\beta}{2}\right)\right) l_6^4 - 156\left(\tan^6\left(\frac{\beta}{2}\right)\right) l_6^4 - 36000 l_6^3 \left(\tan^7\left(\frac{\beta}{2}\right)\right)$$
$$-81000000\left(\tan^8\left(\frac{\beta}{2}\right)\right) l_6^2 + 241\left(\tan^4\left(\frac{\beta}{2}\right)\right) l_6^4$$
$$+42000\left(\tan^5\left(\frac{\beta}{2}\right)\right) l_6^3 + 220500000\left(\tan^6\left(\frac{\beta}{2}\right)\right) l_6^2$$
$$-13500000000 l_6 \left(\tan^7\left(\frac{\beta}{2}\right)\right) + 5062500000000\left(\tan^8\left(\frac{\beta}{2}\right)\right)$$
$$-156\left(\tan^2\left(\frac{\beta}{2}\right)\right) l_6^4 + 42000\left(\tan^3\left(\frac{\beta}{2}\right)\right) l_6^3$$
$$-225000000 l_6^2 \left(\tan^4\left(\frac{\beta}{2}\right)\right) + 67500000000\left(\tan^5\left(\frac{\beta}{2}\right)\right) l_6$$
$$-60750000000000\left(\tan^6\left(\frac{\beta}{2}\right)\right) + 36 l_6^4 - 36000 l_6^3 \tan\left(\frac{\beta}{2}\right)$$
$$+220500000 l_6^2 \left(\tan^2\left(\frac{\beta}{2}\right)\right) + 67500000000\left(\tan^3\left(\frac{\beta}{2}\right)\right) l_6$$
$$+192375000000000\left(\tan^4\left(\frac{\beta}{2}\right)\right) - 81000000 l_6^2$$
$$-13500000000 l_6 \tan\left(\frac{\beta}{2}\right) - 60750000000000\left(\tan^2\left(\frac{\beta}{2}\right)\right)$$
$$+5062500000000 = 0$$

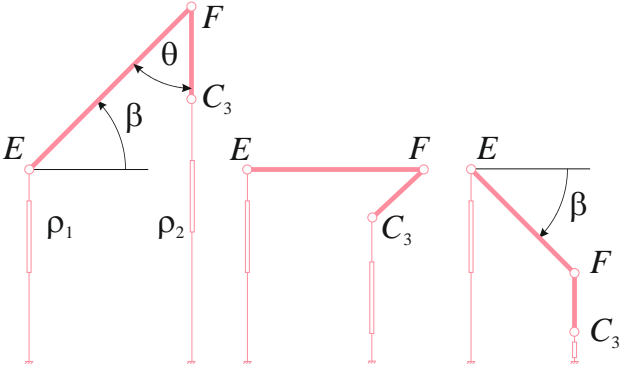

FIGURE 8: POSTURE OF THE MOBILE PLATFORM FOR B= 45°, B= 0° AND B= -45°

From Figure 9, critical point analysis done $\beta$ and $l_6$, which is the minimum length value able to produce the motion
$$\beta = 0.0854, l_6 = 255.885 \quad (23)$$
These values are found by analysing the red regions: $l_6 = 255.885$ is the minimum value where $\beta$ can vary from -45° to +45°.

For the next step, we set $l_6 = 256$. So, now we know the position of $E$ and $C_3$ when $\beta = [-45°, 0, 45°]$.

### 2) Design of the loop I

For a motion of β from 0 to -45°, the vertical displacement of the leg is equal to $2a$. If we set $\beta_{min} = 1/10$ and $\beta_{min} = \pi/2 - 1/10$, we have for the first leg
$$2a = 2l_2 \cos(1/10) - 2l_2 \sin(1/10) \quad (24)$$
$$l_2 = \frac{a}{\cos(1/10) - \sin(1/10)} \quad (25)$$
And for the second one
$$2a = l_4 \cos(1/10) - l_4 \sin(1/10) \quad (26)$$
$$l_4 = \frac{2a}{\cos(1/10) - \sin(1/10)} \quad (27)$$

When $a = 300$, we have $l_2 = 335$ and $l_4 = 670$. The minimal distance between $B_1$ and $D$ is 67. Figure 10 and Figure 11 depict the two leg in the folded and extended positions.

### 3) Design of the loop II

Although the vertical displacements are comparable for this leg, it is necessary to avoid the parallelogram singularity. To address this issue, an offset is introduced, as depicted in Figure 12 [25], as the stiffness of the parallelogram is proportional to its area. This offset does not affect the leg's kinematics, but it enables us to avoid the parallelogram's singularity [21, 25].

Thus, the position of the prismatic joint $A_3$ depends on distance between $B_3$ and $C_3$ along the z-axis and the length $l_6$.

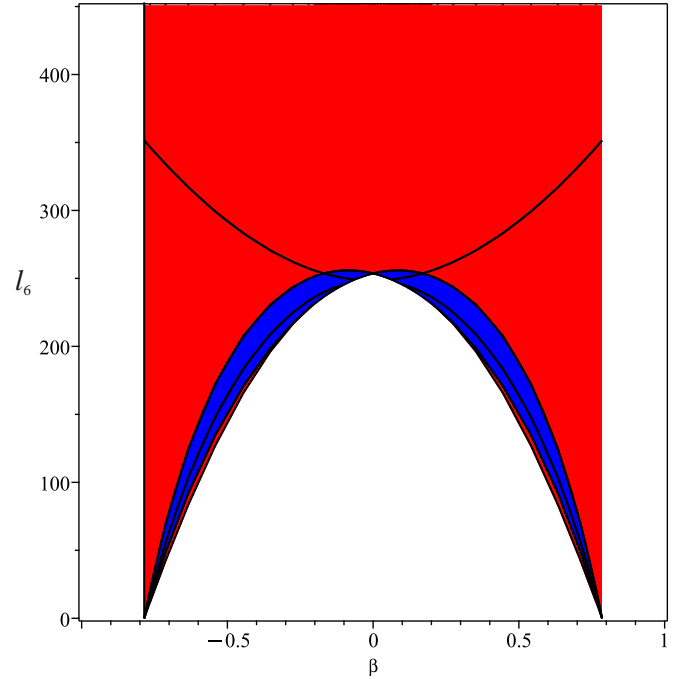

FIGURE 9: CYLINDRICAL ALGEBRAIC DECOMPOSITION OF THE PARAMETER SPACE WITH THE INEQUALITIES

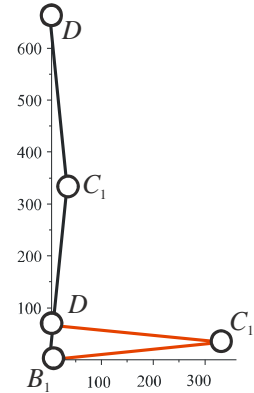
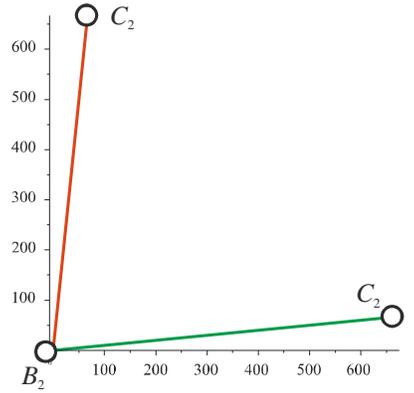

FIGURE 10: FIRST LEG OF THE LOOP I

FIGURE 11: SECOND LEG OF THE LOOP I

### 4) Summary for dimensioning

The decoupling architecture allows us to design the robot for a given Cartesian workspace with one rotation and two translations. Table 3 summarizes the design parameters. The motions of the prismatic joint to move β from -45° to -45° are $y_1 = 0$, $y_2 = 593.5$ and $y_3 = 611.5$. The parameters $l_3$ and $l_8$ are no influence on the kinematic properties but can change the stiffness of the robot.

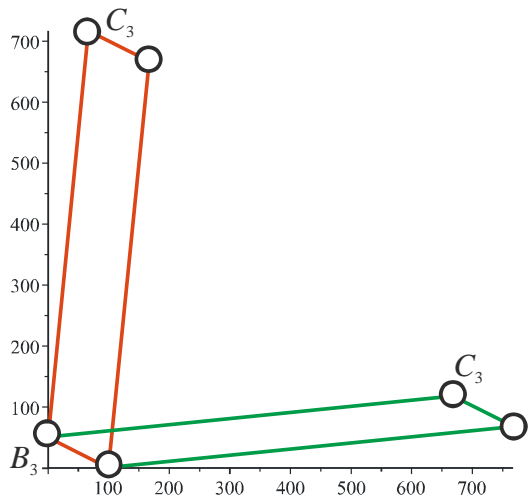

FIGURE 12 : LEG OF THE LOOP II WITH A SMALL OFFSET TO AVOID THE PARALLELOGRAM SINGULARITY

TABLE 3: SIZES OF THE PARALLEL ROBOT

| $a$ | $l_0$ | $l_1$ | $l_2$ | $l_3$ | $l_4$ | $l_5$ | $l_6$ | $l_7$ | $l_8$ |
|---|---|---|---|---|---|---|---|---|---|
| 300 | 10 | $l_0+l_6$ | 335 | 160 | 670 | $l_4$ | 256 | $2a\sqrt{2}$ | 100 |

## 7. CONCLUSIONS

In this article, we have designed a novel 2T1R PM based on the topology design theory of PM using POC. The PM serves as an intelligent sorting and delivery platform for express transfer stations, and is composed of prismatic and rotary pairs, making it simple to manufacture and assemble. The PM's topological characteristics, including DOF and coupling degree, have been calculated, indicating zero coupling degree and partial input-output decoupling. This characteristic greatly benefits the PM's trajectory planning and motion control.

Using the topological characteristic kinematic modeling principle, we have solved the direct and inverse kinematic problem of the PM. Singular analysis has been carried out using the Jacobian matrix, enabling the identification of singular positions and their corresponding configurations. Furthermore, we have presented a case study demonstrating the kinematic behavior of the PM and sizing the mechanism according to a user requirement, which yielded a set of design parameters fulfilling the need.

Overall, our results demonstrate the effectiveness and versatility of the PM design method based on POC, which can be applied to various applications in the field of robotics.

## ACKNOWLEDGEMENTS

The authors gratefully acknowledge the support received from the National Natural Science Foundation of China (No.51975062) and Technology R&D Project of Jiangsu (Prospect and Key Technology for Industry) (No.BE2021016–2).

## REFERENCES

[1] Shen Huiping, Zhou Jinbo, You Jingjing, Yang Tingli., 2T1R parallel mechanism with Analytic Positive Position solutions and Its Kinematic Performance Based Optimization . Transactions of the Chinese Society for Agricultural Machinery, 2020, 51(01):398-409.

[2] Quennouelle, C.,Gosselin, C., Kinematostatic modeling of compliant parallel mechanisms: application to a 3-PRR Mechanism, the Tripteron, Meccanica: Journal of the Italian Association of Theoretical and Applied Mechanics, 2011, 46(1).

[3] Falezza Fabio, Vesentini Federico, Di Flumeri Alessandro, Leopardi Luca, Fiori Gianni, Mistrorigo Gianfrancesco, Muradore Riccardo., A novel inverse dynamic model for 3-DoF delta robots. Mechatronics, 2022, 83.

[4] Farshid Asadi, Ali Heydari., Analytical dynamic modeling of Delta robot with experimental verification, Proceedings of the Institution of Mechanical Engineers, Part K: Journal of Multi-body Dynamics, 2020, 234(3).

[5] Kim Seong Il, Hong Jun Ho, Shin Dongwon, Modeling and motion-control for a Light-weight Delta Robot. The Korean Society of Manufacturing Process Engineers, 2018, 17(3).

[6] Chen Guilan, Cao Yi, Wang Qiang, Workspace Resolution and Dimensional Synthesis on Tsai Allotype DELTA parallel mechanism, Machine Building and Automation, 2016, 45(05):8-12.

[7] Zhang D, Zheng Y, Wei L, et al., Type Synthesis of 2T1R Planar Parallel Mechanisms and Their Modeling Development Applications, IEEE Access, 2021, pp (99):1-1.

[8] Qu S, Li R, Bai S., Type Synthesis of 2T1R Decoupled Parallel Mechanisms Based on Lie Groups and Screw Theory, Mathematical Problems in Engineering, 2017, 2017, pp. 1-11.

[9] Li B, Chen Y, Deng Z, et al., Conceptual design and analysis of the 2T1R mechanism for a cooking robot, Robotics & Autonomous Systems, 2011, 59(2):74-83.

[10] D Ding, Zhang Y, Xin W, et al., Kinematic Analysis and Simulation of a Novel 2T1R Parallel Mechanism, 2016 4th International Conference on Machinery, Materials and Computing Technology. Atlantis Press, 2016.

[11] Shen Huiping, Tang Yao, Wu Guanglei, et al., Design and analysis of a class of two-limb non-parasitic 2T1R parallel mechanism with decoupled motion and symbolic forward position solution-influence of optimal arrangement of limbs onto the kinematics, dynamics and stiffness, Mechanism and Machine Theory, 2022, 172.

[12] Yongquan Li, Yang Zhang, Lijie Zhang., A New Method for Type Synthesis of 2R1T and 2T1R 3-DOF Redundant Actuated Parallel Mechanisms with Closed Loop Units, Chinese Journal of Mechanical Engineering, 2020, 33(06):144-167.

[13] Huang, K.W., Shen, H.P., Li, J., Zhu, Z.C., Yang, T.L., Topological Design and Dynamics Modeling of a Spatial 2T1R Parallel Mechanism with Partially Motion Decoupling and Symbolic Forward Kinematics. China Mechanical Engineering, 2022, 33 (2): 160-169.

[14] Yang, T.L., Robot mechanism topology design, Science Press, 2012.

[15] T.L. Yang, A.X. Liu, H.P. Shen, et al., Topology Design

of Robot Mechanisms, Springer, 2018.

[16] Shen Huiping, Topological Characterization Kinematics of Robot Mechanisms, Beijing: Higher Education Press, 2021.

[17] Melet J-P, Parallel Robots, Vol. 128. Springer Science & Business Media, 2006.

[18] Jha R, Chablat D, Baron L, Rouillier F., Moroz G., Workspace, joint space and singularities of a family of delta-like robot, Mechanism and Machine Theory, 2018, 127:73-95.

[19] Chablat, D., Wenger, P., Merlet, J., Workspace analysis of the Orthoglide using interval analysis, In Advances in Robot Kinematics, Springer, Dordrecht, June 24-28, pp. 397-406, 2002.

[20] Shen, H., Chablat, D., Zen, B., Li, J., Wu, G., Yang, T., A Translational Three-Degrees-of-Freedom Parallel Mechanism With Partial Motion Decoupling and Analytic Direct Kinematics, Journal of Mechanisms and Robotics, April 2020, Vol.12/021112-1-7

[21] Müller A., Zlatanov D., Singular Configurations of Mechanisms and Manipulators, Switzerland: Springer, 2019.

[22] Wu G, Shen H. Parallel PnP Robots-Parametric Modeling, Performance Evaluation and Design Optimization, Springer Nature, 2020.

[23] Moroz, Guillaume, Chablat Damien, Rouillier Fabrice, Moroz Guillaume, Cusp Points in the Parameter Space of RPR-2PRR Parallel Manipulators, New Trends in Mechanism Science: Analysis and Design. Springer Netherlands, 2010.

[24] Chablat Damien, Moroz Guillaume, Rouillier, Fabrice, Wenger, Philippe, Using Maple to analyse parallel robots. In Maple in Mathematics Education and Research: Third Maple Conference, MC 2019, Waterloo, Canada, October 15–17, 2019, Springer International Publishing; 2020, pp. 50-64.

[25] Majou, Félix, Philippe Wenger, Damien Chablat, Design of a three-axis machine tool for high speed machining, Proceedings of the 4th International Conference on Integrated Design in Manufacturing and Mechanical Engineering Clermont-Ferrand, France, 2002.